\documentclass[journal]{IEEEtran}
\usepackage{amsmath,amsfonts,amssymb}
\usepackage{algorithmic}
\usepackage{algorithm}
\usepackage{array}
\usepackage[caption=false,font=normalsize,labelfont=sf,textfont=sf]{subfig}
\usepackage{textcomp}
\usepackage{stfloats}
\usepackage{url}
\usepackage{verbatim}
\usepackage{graphicx}
\usepackage{cite}
\usepackage{soul}
\usepackage{xcolor}
\sethlcolor{yellow}

\usepackage{booktabs}   
\usepackage{multirow}   
\usepackage{threeparttable}

\usepackage{hyperref}

\hypersetup{
    pdfauthor={Anonymous},
    pdftitle={ImagineUAV}
}

\begin{document}
	
\title{ImagineUAV: Aerial Vision-Language Navigation via World-Action Modeling and Kinodynamic Planning}

\author{
    \vskip 1em
    Xuchen Liu, Jiawei Huang, Shihao Xia, Bingxi Liu, Jinqiang Cui and Jiankun Yang
    \thanks{Xuchen Liu, Jinqiang Cui and Jiankun Yang are with Pengcheng Laboratory, Shenzhen, Guangdong, China (e-mail: \{liuxch,cuijq,jiankun\}@pcl.ac.cn). }
    \thanks{Jiawei Huang is with the School of Computer Science and Cyber Engineering, Guangzhou University, Guangzhou, Guangdong, China (e-mail: 2112506155@e.gzhu.edu.cn).}
    \thanks{Shihao Xia and Bingxi Liu are with Southern University of Science and Technology, Shenzhen, Guangdong, China (e-mail: 12543041@mail.sustech.edu.cn, binuxliu@gmail.com).}
}




\maketitle

\begin{abstract}
	Vision-language navigation (VLN) for UAVs demands grounding free-form instructions into 6-DoF flight under partial observability. While Vision-Language-Action (VLA) models excel at semantic reasoning, they suffer from brittleness due to geometric inconsistency and dynamics mismatch. To address this, we propose ImagineUAV, an imagination-driven framework leveraging cascaded world-action modeling. Instead of direct regression, ImagineUAV employs a latent video diffusion model to generate instruction-conditioned future observations—explicitly ``imagining'' environmental evolution—from which 6-DoF motions are inferred via an action extractor. A kinodynamic planner then refines these estimates into collision-free trajectories. Additionally, a step-distilled inference pipeline ensures real-time execution. With only 1.3B parameters, ImagineUAV outperforms prior VLN and VLA baselines on benchmarks and real-world flights, validating the practicality of imagination-driven aerial navigation. Video demo: \url{https://www.youtube.com/watch?v=Ng1alP0yhc0}.
\end{abstract}


\section{Introduction}
\IEEEPARstart{V}{ision}-language navigation (VLN) investigates how an embodied agent executes free-form natural-language instructions in unseen environments. Unlike conventional autonomous navigation, VLN is fundamentally instruction-centric and perception-driven: the agent must continuously align linguistic intent with streaming egocentric observations and infer actions under partial observability~\cite{wu2024vision}. In aerial settings, the difficulty is further amplified by six-degree-of-freedom (6-DoF) motion, rapid viewpoint variation, payload-constrained sensing, and strict safety margins in cluttered 3D spaces. Small grounding or geometric estimation errors can escalate quickly into unstable or unsafe flight behavior, especially when operators issue high-level commands (e.g., ``fly through the corridor and rise above the shelf'') instead of dense waypoint trajectories. 

With the rapid rise of language-based foundation models, especially large language models (LLMs) and vision-language models (VLMs), an increasing number of VLN methods have adopted them as high-level planners, yielding notable gains in instruction understanding, semantic grounding, and subgoal decomposition. However, this planning strength fails to transfer to low-level UAV control: 6-DoF flight demands geometric consistency and dynamics awareness that transcend semantic reasoning. VLA models~\cite{zhang2024navid} attempt to bridge this by introducing action tokens into the transformer backbone. However, the LLM/VLM backbone remains optimized for semantic prediction rather than embodied dynamics, making learned action heads brittle under out-of-domain shifts~\cite{kim2024openvla}. Hence, a core challenge persists: how to produce executable low-level commands from language-level intent while maintaining geometric consistency under real-world UAV dynamics.

In parallel, video-generation-based world models have progressed rapidly, showing a strong ability to predict temporally coherent future observations conditioned on context or action~\cite{wan2025wan}. These advances motivate a different route: instead of forcing language models to output controls directly, one can ask a world model to \emph{imagine what the agent would see if it follows the instruction}~\cite{zhao2025airscape}. If imagination is sufficiently realistic and instruction-aligned, the generated video implicitly contains a future route, obstacle interaction cues, and motion scale. A geometric backend can then extract executable trajectories from this visual future. 

\begin{figure*}[!t]
	\centering
	\includegraphics[width=0.9\textwidth]{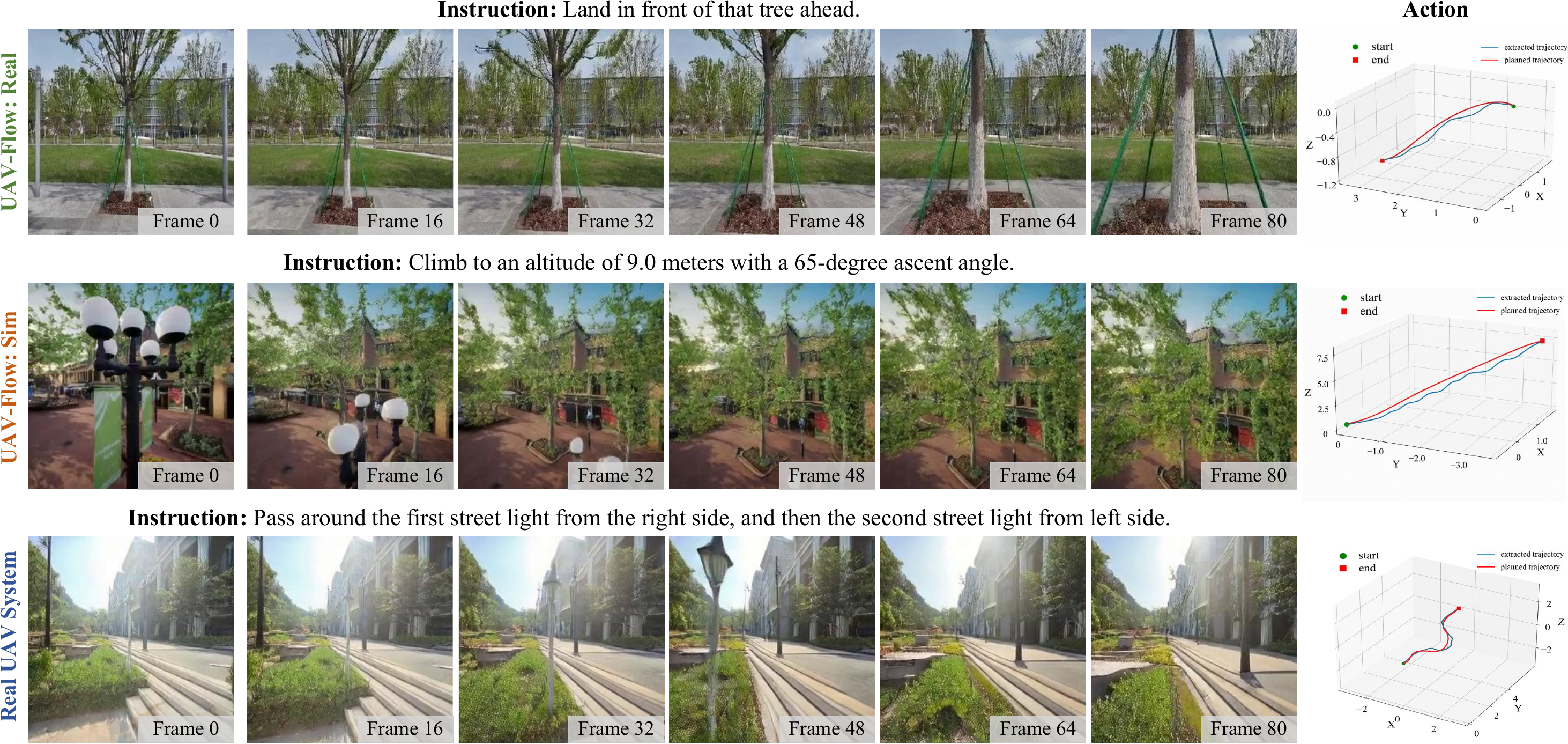}
	\caption{Imagination-guided UAV navigation across diverse environments. Each row illustrates the instruction-conditioned generated observations alongside the corresponding flight execution results. The top and middle rows present UAV-Flow~\cite{wang2025uav} cases, while the bottom row shows a real UAV system case. }
	\label{fig:intro_case}
\end{figure*}

Based on this insight, we propose \textit{ImagineUAV}, an imagination-driven UAV vision-language navigation framework composed of three tightly coupled modules: (i) an instruction-conditioned world model that predicts future egocentric observations over a planning horizon; (ii) a geometry-aware action module implemented through learning-based visual odometry that extracts relative 6-DoF poses from imagined frames; and (iii) a kinodynamic planner that converts the extracted poses into collision-free, dynamically feasible trajectories. Specifically, through instruction-conditioned training, we enable the video generation model to accept VLN instructions and produce embodied future observation sequences covering diverse 6-DoF aerial motion patterns, which serve as a bridge between semantics and geometry. Language contributes task intent, and the world model materializes intent into controllable pixel-level futures. Unlike videos induced by real physical motion, imagined videos can exhibit spatio-temporal inconsistencies (e.g., object drifting and local temporal jitter); therefore, we carefully design a learning-based visual-odometry action extractor that is robust to such artifacts and still recovers reliable pose series at execution time. Since the raw poses may violate dynamical constraints and lack explicit collision checking, a kinodynamic planner refines them into smooth, collision-free, dynamically feasible trajectories for reliable flight execution. Finally, we deploy the system on a real UAV platform, where step-distilled inference provides practical planning latency and the onboard flight stack executes the generated waypoint references. Experiments validate the feasibility of our method, as illustrated in Fig.~\ref{fig:intro_case}. 

The contributions of this paper are summarized as follows:
\begin{itemize}
\item We propose an instruction-conditioned latent video diffusion world model for 6-DoF aerial VLN, which explicitly imagines future egocentric observations conditioned on language instructions and grounds motion intent through masked camera-control training.

\item We design a spatio-temporal action extractor that robustly infers executable relative motions from imagined videos. A downstream kinodynamic planner further converts these estimates into collision-free and dynamically feasible trajectories, mitigating the brittleness of direct action execution.

\item We integrate the proposed modules into a UAV navigation system with step-distilled inference for low-latency world-model generation. Extensive simulation experiments on the UAV-Flow benchmark~\cite{wang2025uav} demonstrate state-of-the-art performance (70.9\% success rate), while real-world flight tests consistently validate the effectiveness and practicality of the proposed framework.
\end{itemize}

\section{Related Work}

\subsection{Vision-Language Navigation and Action Learning}
VLN has progressed from discrete graph-based benchmarks~\cite{anderson2018vision} to continuous-control settings~\cite{krantz2020beyond} that require agents to execute low-level motion commands under noisy egocentric perception, better reflecting real-world robotic constraints. More recently, foundation-model-based approaches treat VLN as a high-level planning problem~\cite{ahn2022can}, leveraging LLMs and VLMs for instruction decomposition while delegating low-level control to separate modules. A popular paradigm, VLA models~\cite{brohan2023rt1,zitkovich2023rt,sapkota2025vision}, unifies perception, reasoning, and control by co-fine-tuning multimodal foundation models on large-scale robotic trajectories~\cite{kim2024openvla}, predicting end-to-end action tokens directly from visual and linguistic inputs. In aerial settings, VLA frameworks~\cite{gao2025openfly,wang2025uav} have been proposed to regress 6-DoF control commands from raw pixels. However, direct action regression remains fundamentally limited by the lack of explicit future-state reasoning: without anticipating how the scene evolves, VLA backbones—optimized primarily for semantic prediction—struggle with geometric inconsistency, dynamics mismatch, and compounding errors over long horizons. This motivates an alternative paradigm that reasons over imagined future observations before committing to executable actions.

\subsection{World Action Models}
Recent advances in generative video modeling have enabled powerful predictive models of future observations.
By combining these advances with the objective of modeling environment dynamics for planning and control~\cite{hafner2025mastering}, World Action Models (WAMs)~\cite{zhang2025world} jointly model future observations and actions within a unified generative framework. These approaches have diverged along three complementary routes. Latent-space methods~\cite{ye2026world} predict compressed future states in a learned embedding and decode actions directly, offering computationally efficient planning but without explicit visual output. Pixel-level methods~\cite{bar2025navigation,zhao2025airscape} generate full future visual observations, preserving scene structure and motion scale for human-interpretable inspection. Hybrid approaches such as MoWM~\cite{shang2025mowm} combine both paradigms, reporting that latent representations better capture dynamics while pixel features preserve geometric details necessary for precise action decoding. Despite this progress, existing WAM research has primarily focused on manipulation, policy learning, and direct action decoding in limited action spaces. The integration of instruction-conditioned visual imagination with executable action generation and motion planning remains largely unexplored for aerial vision-language navigation. Moreover, existing approaches~\cite{huang2026navdreamer} often rely on large external generative models and incur substantial inference costs without domain-specific adaptation.

\section{Method}
\begin{figure}[!t]
	\centering
	\includegraphics[width=\linewidth]{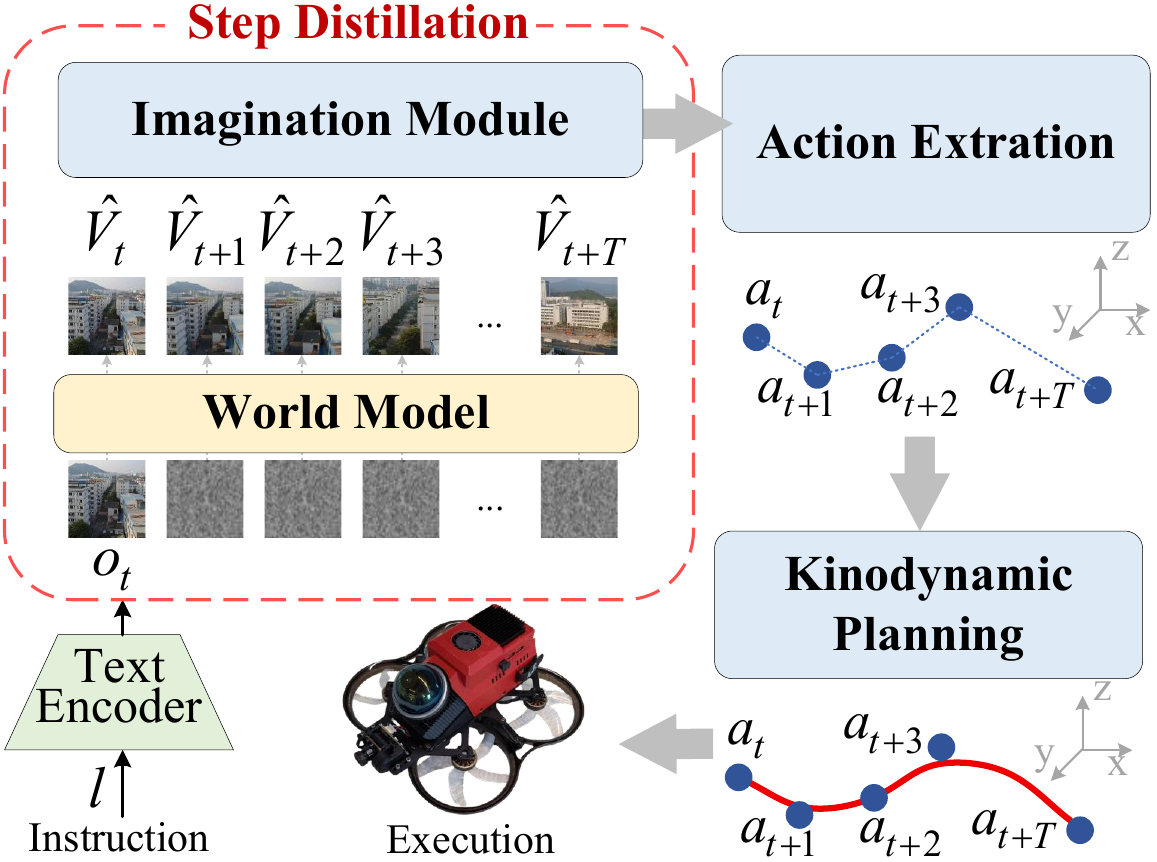}
	\caption{\textbf{Framework Overview:} The pipeline is composed of three primary stages: (i) the \textbf{Instruction-Conditioned Imagination Module}, where a Text Encoder processes the VLN instruction $l$ to guide a Diffusion Transformer in predicting a future egocentric video rollout $\{\hat{V}_{t:t+T}\}$ based on the current context observation $o_t$; (ii) the \textbf{Action Extraction Module}, which employs a learned visual-odometry action extractor to infer executable relative motions $\{a_{t:t+T}\}$ directly from the imagined visual frames; and (iii) the \textbf{Kinodynamic Planner}, which refines motions into smooth, collision-free, dynamically feasible trajectories via gradient-based optimization. Notably, the Imagination Module is further accelerated via a step distillation strategy to preserve generation fidelity while reducing inference steps.}
	\label{fig:method}
\end{figure}
\subsection{Problem Formulation}
We consider the VLN task as a partially observable problem. Given a natural-language instruction $l$ and initial state $s_0$, the UAV must navigate to a goal region $g$ using only egocentric observations $o_t$, without access to a global map. At each time step $t$, the agent observes the environment and must produce a control action that drives it toward the goal while adhering to the given instruction.

\subsection{System Overview}\label{sec:overview}
ImagineUAV decomposes the VLN task into three cascaded stages (Fig.~\ref{fig:method}): an imagination module $\mathcal{W}$, a visual-odometry extractor $D_\psi$, and a kinodynamic planner $\mathcal{K}$. Given observation $o_t$ and instruction $l$, the pipeline proceeds as:
\begin{align}
	\hat{V}_{t:t+T} &= \mathcal{W}(o_t,l), \\
	a_{t:t+T} &= D_\psi(\hat{V}_{t:t+T}), \\
	\tau^* &= \mathcal{K}(\{a_{t+k}\},\,\mathcal{O}),
\end{align}
where $a_{t:t+T}=\{a_{t+k}\}_{k=1}^{T}$ is the sequence of predicted relative motions, each $a_{t+k}=(\Delta \hat{p}_{t+k}, \Delta \hat{r}_{t+k})$ being a relative translation--rotation pair. $\mathcal{O}$ denotes the local obstacle map. The kinodynamic planner $\mathcal{K}$ ensures that $\tau^*$ is smooth, collision-free, and dynamically feasible.

\subsection{Imagination Module based on World Model}
\subsubsection{Backbone and Conditioning}
Our imagination module instantiates an instruction-conditioned video diffusion transformer, designed to model the conditional transition distribution of egocentric aerial observations over a finite horizon. Given the current observation $o_t$ and language instruction $l$, the model generates a future clip $\hat{V}_{t:t+T}=\{\hat{o}_{t+1},\ldots,\hat{o}_{t+T}\}$ through iterative denoising in a compact latent space:
\begin{equation}
	\hat{V}_{t:t+T}\sim p_{\theta}(V\mid o_t,l).
\end{equation}
Concretely, instruction tokens are embedded by a language encoder, while the current observation and temporal context are mapped into visual latent tokens by a video tokenizer. Multi-layer cross-attention and adaptive normalization are then used to inject linguistic intent into spatio-temporal generation at different scales. This design enables semantic controllability at the instruction level while preserving local geometric continuity required by UAV motion.

\subsubsection{Objectives}
The imagination module is trained to generate instruction-consistent and temporally coherent future egocentric observations that are suitable for downstream geometric extraction. Since the module serves as a semantic-to-visual bridge rather than a control policy, its objectives focus on three key properties: (i) fidelity to the observed visual dynamics, (ii) alignment with language-specified motion intent, and (iii) temporal smoothness that supports robust action extraction.

Let $x_0$ denote the ground-truth future-video latent sampled from paired instruction--video sequences, and let $\tau$ denote the diffusion timestep. We adopt a standard forward diffusion process
\begin{equation}
	q(x_{\tau}\mid x_0)=\mathcal{N}\!\left(\sqrt{\bar{\alpha}_{\tau}}x_0,(1-\bar{\alpha}_{\tau})I\right),
\end{equation}
which progressively corrupts the clean latent with Gaussian noise. The imagination backbone is trained to invert this process by predicting the injected noise conditioned on the current observation and the instruction.

Following denoising diffusion probabilistic models, the primary training objective is a conditional noise-prediction loss:
\begin{equation}
	\mathcal{L}_{\mathrm{diff}}=\mathbb{E}_{x_0,\epsilon,\tau,o_t,l}\left[\left\|\epsilon-\epsilon_{\theta}(x_{\tau},\tau,o_t,l)\right\|_2^2\right],
\end{equation}
where $\epsilon\sim\mathcal{N}(0,I)$ and $x_{\tau}=\sqrt{\bar{\alpha}_{\tau}}x_0+\sqrt{1-\bar{\alpha}_{\tau}}\,\epsilon$. Conditioning on $(o_t,l)$ encourages the model to generate futures that are semantically consistent with the instruction, such as forward motion, altitude changes, or turning behaviors, while remaining grounded in the visual context of the current observation.

However, the diffusion objective implicitly emphasizes temporal consistency across frames: abrupt appearance changes, inconsistent camera motion, or unrealistic viewpoint jumps are penalized during training because they increase reconstruction difficulty across diffusion steps. This property is critical for downstream action extraction, which assumes smooth inter-frame motion to recover relative pose reliably.

In practice, the objective is optimized over large-scale instruction-video data collected from both simulated and real UAV platforms. By training on long-horizon egocentric sequences with diverse motion patterns, the model learns a distribution over feasible future observations under aerial dynamics. As a result, the imagination module produces visually plausible and instruction-aligned rollouts that, while not perfectly physically accurate at the pixel level, preserve sufficient spatio-temporal structure for robust trajectory extraction subsequently.

\subsection{Action Extraction and Kinodynamic Planning}
A learning-based action extractor $D_{\psi}$ maps the imagined video clip $\hat{V}_{t:t+T}=\{\hat{o}_{t+1},\ldots,\hat{o}_{t+T}\}$ to the relative motion sequence $a_{t:t+T}$ (defined in Section~\ref{sec:overview}).

To better match the output distribution of the world model, which may differ from real camera videos due to local texture drift and temporal artifacts, we design a learning-based extractor under a Transformer framework. It tokenizes short monocular clips into spatio-temporal visual tokens, uses self-attention to aggregate appearance and motion cues across frames, and regresses relative translation and rotation through lightweight prediction heads.

The extractor $D_{\psi}$ is trained from scratch on video--trajectory pairs $\{(V^{(i)},P^{(i)})\}_{i=1}^{N}$, where $P^{(i)}=\{a^{*(i)}_{t+1},\ldots,a^{*(i)}_{t+T}\}$ denotes the ground-truth relative trajectory, and optimized with supervised pose regression:
\begin{equation}
	\mathcal{L}_{\mathrm{act}}
	=\sum_{k=1}^{T}\left(
	\left\|\Delta\hat{p}_{t+k}-\Delta p^{*}_{t+k}\right\|_1
	+\lambda_r\left\|\Delta\hat{r}_{t+k}-\Delta r^{*}_{t+k}\right\|_1
	\right),
\end{equation}
where $a^{*}_{t+k}=(\Delta p^{*}_{t+k},\Delta r^{*}_{t+k})$ is the supervised relative motion label.

The predicted motions $a_{t:t+T}$ output by $D_{\psi}$ are cumulatively summed into a world-frame reference trajectory, which encodes obstacle-aware semantics from the imagined video but may violate dynamic constraints or encounter unforeseen obstacles. A kinodynamic planner refines this reference into a smooth, collision-free, dynamically feasible trajectory.

Kinodynamic planning is formulated as a gradient-based trajectory optimization~\cite{zhou2020fastplanner}. The trajectory is parameterized as a uniform B-spline with control points $\{\mathbf{q}_i\}_{i=0}^{N_p}$, and the composite objective is
\begin{align}
J &= C_{\mathrm{smooth}} + C_{\mathrm{obs}} + C_{\mathrm{feas}} + C_{\mathrm{ref}} \label{eq:kinodynamic_J}\\
C_{\mathrm{smooth}} &= \textstyle\sum_{i} \|\mathbf{q}_{i+1} - 2\mathbf{q}_i + \mathbf{q}_{i-1}\|^2 \notag\\
C_{\mathrm{obs}} &= \textstyle\sum_{i} \max\!\big(0,\, d_{\min} - d_s(\mathbf{q}_i)\big)^2 \notag\\
C_{\mathrm{feas}} &= \textstyle\sum_{i} \big[\max\!(0,\|\mathbf{v}_i\|-v_{\max})^2 \notag\\
&\quad + \max\!(0,\|\mathbf{a}_i\|-a_{\max})^2\big] \notag\\
C_{\mathrm{ref}} &= \textstyle\sum_{i} d(\mathbf{q}_i)^2 \notag
\end{align}
where $C_{\mathrm{smooth}}$ penalizes the elastic-band energy of consecutive control points, $C_{\mathrm{obs}}$ repels control points from obstacles via a Euclidean signed distance field ($d_s$), and $C_{\mathrm{feas}}$ penalizes velocity and acceleration violations ($\mathbf{v}_i$, $\mathbf{a}_i$ from B-spline derivatives). The reference-deviation cost $C_{\mathrm{ref}}$ anchors the trajectory to the imagined path, with $d(\mathbf{q}_i)$ being the minimum distance from $\mathbf{q}_i$ to the polyline integrated from $\{a_{t+k}\}$. The gradient of each cost is integrated into the L-BFGS optimizer. A reference-guided local goal is selected along the polyline at arc length $s^* + L_{\mathrm{ahead}}$ from the UAV's nearest projection, where $s^*$ denotes the projected arc-length coordinate, anchoring each receding-horizon replan to the imagined route.

\subsection{Step-Distilled Deployment}
The generative world model is too computationally demanding for the UAV to execute with acceptable latency. Therefore, our deployment design focuses on accelerating the imagination module through step distillation.

We adopt step distillation as the core acceleration strategy. Following Distribution Matching Distillation (DMD), we formulate distillation as a conditional distribution matching problem. Let $x$ denote a clean future-video latent and $c=(o_t,l)$ the conditioning context. A sparse-step student generator $G_{\theta_{\mathrm{stu}}}(\eta,c)$ maps an initial Gaussian latent to a generated clean latent, and is optimized to minimize the KL divergence $D_{\mathrm{KL}}(p_{\mathrm{stu}}\|p_{\mathrm{tea}})$ between the student and teacher distributions. Since both densities are intractable, DMD uses the gradient of this divergence, estimated by the difference between a fixed teacher score model $s_{\mathrm{tea}}(x,c)=\nabla_x\log p_{\mathrm{tea}}(x|c)$ and an auxiliary fake-score model $s_{\mathrm{stu}}(x,c)=\nabla_x\log p_{\mathrm{stu}}(x|c)$ that tracks the current student distribution:
\begin{equation}
	\nabla_{\theta_{\mathrm{stu}}}D_{\mathrm{KL}}
	=\mathbb{E}_{\eta,c}\left[
	-\left(s_{\mathrm{tea}}(x,c)-s_{\mathrm{stu}}(x,c)\right)
	\frac{\partial G_{\theta_{\mathrm{stu}}}(\eta,c)}{\partial\theta_{\mathrm{stu}}}
	\right],
\end{equation}
so that the student's few-step outputs move toward the teacher distribution under the same observation--instruction condition.

The student generator uses an $m$-step denoising schedule, reducing the original 40--50 denoising evaluations to ten DiT evaluations. We further distill classifier-free guidance into the student by using the guided teacher noise prediction:
\begin{equation}
\begin{aligned}
	\epsilon^{\mathrm{cfg}}_{\mathrm{tea}}(x_{\tau},\tau,o_t,l)
	=&\;\epsilon_{\theta_{\mathrm{tea}}}(x_{\tau},\tau,o_t,\emptyset) \\
	&+w\Big(
	\epsilon_{\theta_{\mathrm{tea}}}(x_{\tau},\tau,o_t,l)
	-\epsilon_{\theta_{\mathrm{tea}}}(x_{\tau},\tau,o_t,\emptyset)
	\Big),
\end{aligned}
\end{equation}
where $\emptyset$ denotes the dropped language condition while the current observation $o_t$ is retained, and $w$ is the guidance scale. Consequently, inference requires only ten conditional DiT evaluations and disables CFG, substantially reducing planning latency for navigation.

\section{Experiments}
\subsection{Experimental Setup}

\textbf{Dataset.} We conduct our experiments using the UAV-Flow~\cite{wang2025uav} dataset, which focuses on the Flow paradigm—characterized by short-range, fine-grained, and dynamically realistic execution—to intuitively validate the spatial understanding and predictive consistency of our world model in VLN. The dataset comprises 30k real-world trajectories and 10k simulation trajectories across 10 major motion types, primarily within a 20-meter range. We utilize all available real and simulation sequences for joint training and validation, while systematic evaluation is performed on the official simulation test set. 

\textbf{Imagination Module.} The backbone of our imagination module is Wan2.1-Fun-V1.1-1.3B-Control-Camera~\cite{he2024cameractrl}. Compared to the vanilla Wan-1.3B model~\cite{wan2025wan}, this version supports image-conditioning, allowing the model to take the current observation $o_t$ as a structural prior. To prevent the model from relying on predefined camera trajectories and force it to ground motion solely on language instructions, we mask the camera-control latents during training and testing. Training is conducted on 8$\times$ NVIDIA PRO 6000 (96GB) GPUs. We use the AdamW optimizer with a learning rate of $1\times10^{-5}$ and a weight decay of $0.01$. The model is trained for 2 epochs and the video resolution is set to $256 \times 256$. In the step distillation, $m$ is set to 10.

\textbf{Action Extraction.} We train the neural action extractor $D_{\psi}$ from scratch using paired monocular UAV videos and ground-truth relative trajectories collected from both UE5-based simulation environments and real-world flights equipped with external positioning systems. The extractor takes short egocentric clips as input and predicts relative 6-DoF motions with confidence scores. During testing, it processes imagined rollouts in a sliding-window manner, composes the predicted relative motions, and smooths low-confidence segments before sending waypoint references to the low-level controller.

\begin{figure}[!t]
	\centering
	\includegraphics[width=\linewidth]{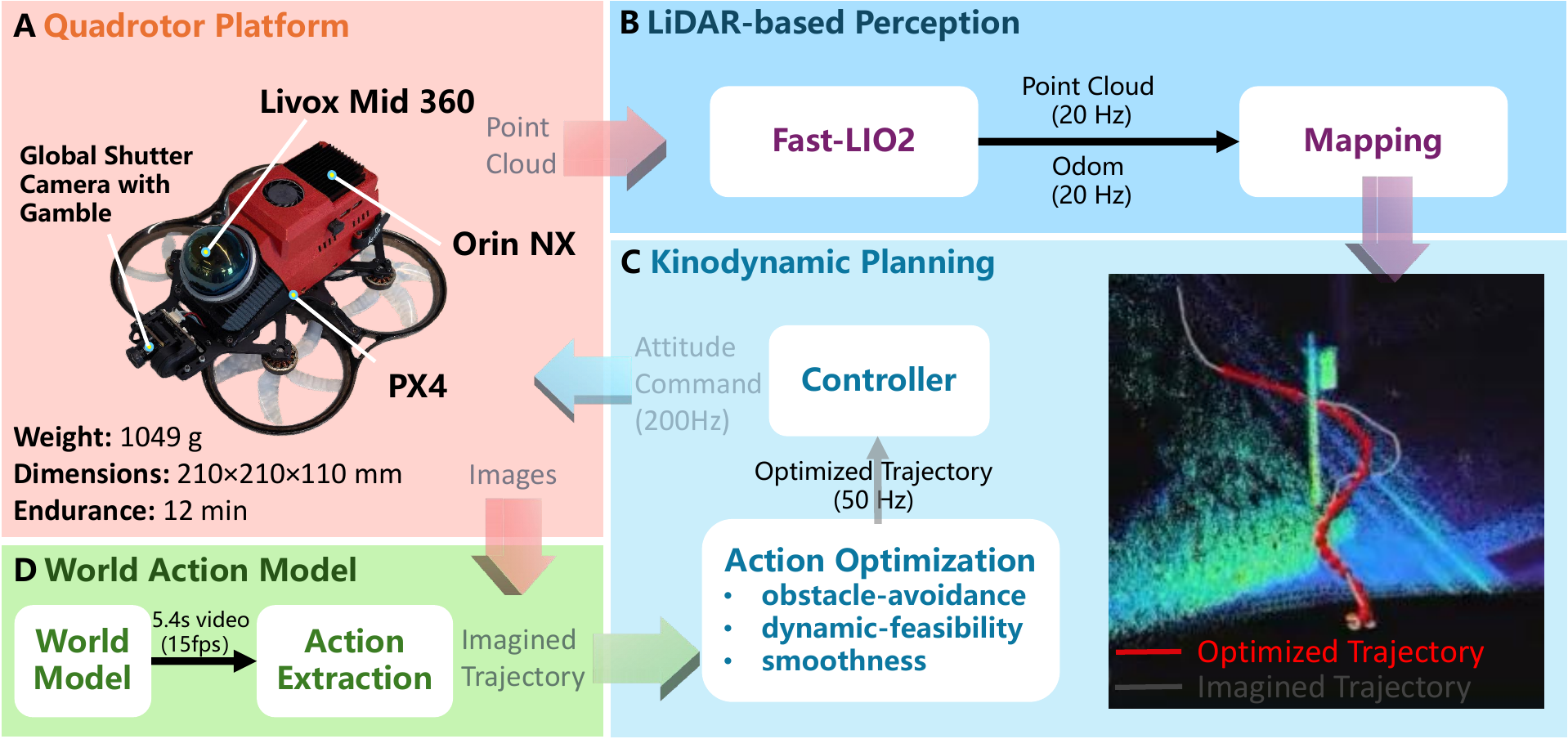}
	\caption{
        \textbf{Real-world deployment:} 
        (A) The quadrotor platform integrates an onboard camera, LiDAR, PX4 flight controller, and Orin NX computer for fully onboard perception and control. 
        (B) LiDAR-based perception employs FAST-LIO2 for real-time odometry estimation and mapping. 
        (C) A kinodynamic planner optimizes imagined trajectories into dynamically feasible and collision-free flight paths. 
        (D) The proposed World Action Model generates future observations and extracts executable 6-DoF motions for downstream planning.
}
	\label{fig:deployment}
\end{figure}

\textbf{Real-world deployment.}  We deploy on a custom 3.5-inch FPV quadrotor\footnote{{Hardware Detials: \url{https://qgduan.github.io/Spikive-Astro/}}}. integrating an onboard Orin NX computer, LiDAR, monocular camera, and PX4 flight stack (Fig.~\ref{fig:deployment}A). FAST-LIO2 provides real-time odometry and mapping onboard (Fig.~\ref{fig:deployment}B), while the kinodynamic planner runs on Orin NX (Fig.~\ref{fig:deployment}C). An external edge unit with an NVIDIA RTX PRO 6000 GPU executes the world model to generate $5.4s$ imagined rollouts (Fig.~\ref{fig:deployment}D).

\subsection{Quantitative Results}

\begin{figure*}[htbp]
	\centering
	\includegraphics[width=0.97\textwidth]{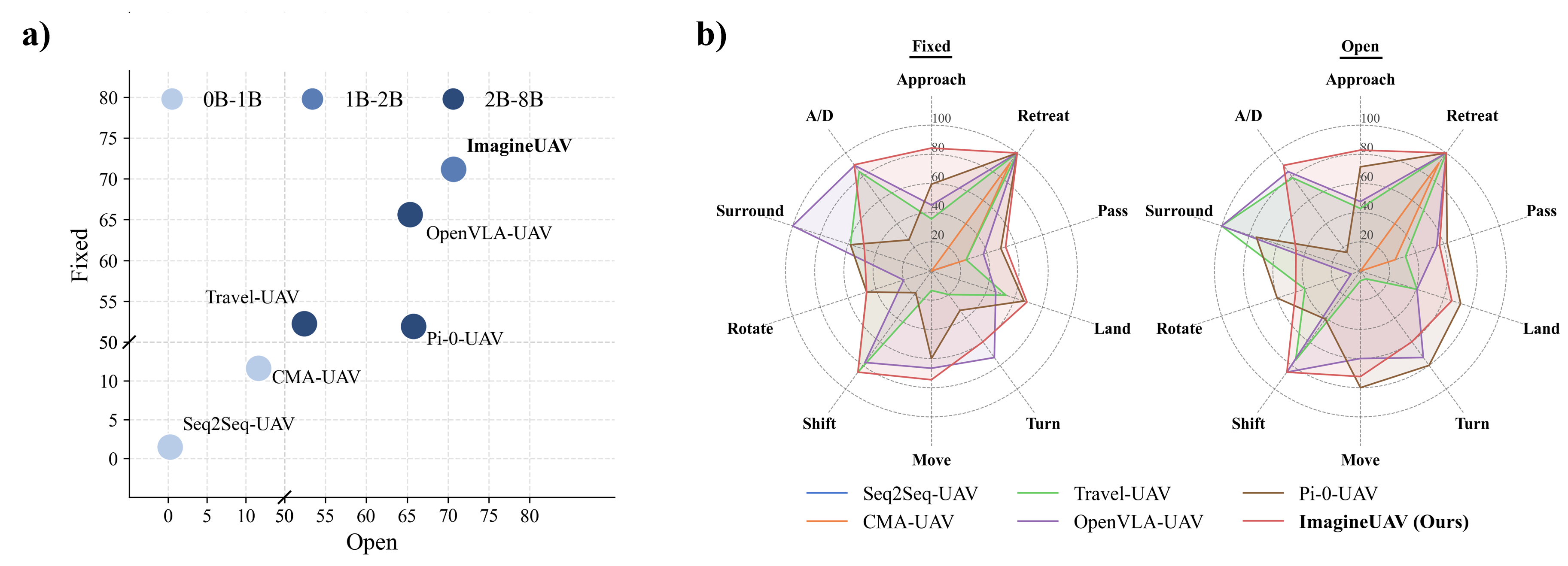}
	\caption{Success rates (\%) on the UAV-Flow-Sim test set. \textbf{Fixed} and \textbf{Open} denote fixed-template and open-vocabulary command sets, respectively. Tasks include: \textit{Approach} (target proximity), \textit{Retreat} (moving away), \textit{Pass} (obstacle traversal), \textit{Land} (precision landing), \textit{Turn} (heading toward target), \textit{Move/Shift} (metric/geometric translation), \textit{Rotate} (in-place angular adjustment), \textit{Surround} (orbital flight), and \textit{A/D} (Ascend/Descend, vertical altitude control).}
	\label{fig:transposed_results}
\end{figure*}
\vspace{1mm}

Fig.~\ref{fig:transposed_results} reports the success rates of different approaches on the UAV-Flow-Sim benchmark across ten motion categories. ImagineUAV achieves the highest average success rate, 70.9\% (fixed 71.19\% and open-vocabulary 70.65\%), outperforming both traditional VLN methods and recent VLA-based baselines.

\textbf{Comparison with Baseline Paradigms.}
Traditional VLN models such as \textbf{Seq2Seq-UAV} and \textbf{CMA-UAV} perform poorly in the continuous Flow setting, with average success rates below 15\%. Their discrete-action formulations and limited geometric reasoning result in severe trajectory drift and inaccurate termination. VLA-based approaches, including \textbf{OpenVLA-UAV} and \textbf{Pi-0-UAV}, substantially improve overall performance by unifying perception and action, achieving average success rates in the mid-60\% range. In contrast, ImagineUAV further improves the average success rate by explicitly decoupling semantic foresight from geometric execution, yielding more stable and instruction-faithful trajectories.

\textbf{Model Scale Advantage.}
The size comparison in Fig.~\ref{fig:transposed_results}(a) further highlights the efficiency of ImagineUAV. While recent VLA baselines rely on large foundation backbones, typically at the 3B--7B scale, ImagineUAV achieves stronger navigation performance with a compact 1.3B-scale parameterization. This indicates that, for UAV navigation, explicitly modeling visual imagination and extracting geometry-aware actions can provide a more favorable performance--scale trade-off than directly scaling end-to-end VLA policies. Such compactness is beneficial to practical UAV deployment.

\textbf{Geometric Foresight and Kinodynamic Planning.}
The benefits of imagination-driven planning are most evident in tasks that require accurate anticipation of spatial extent and motion scale. ImagineUAV achieves strong performance in \textit{Approach}, \textit{Move}, \textit{Shift}, and vertical \textit{A/D} tasks, reaching 84.31\%, 74.57\%, 85.71\%, and 90.02\% under fixed instructions, respectively. These results suggest that the imagined visual future provides meaningful geometric foresight, while the kinodynamic planner bridges high-level intention prediction and executable flight control, particularly benefiting interaction-sensitive tasks such as \textit{Pass} and \textit{Land}.

\textbf{Limitations and Failure Analysis.}
Despite the overall superiority, ImagineUAV remains less effective in yaw-dominant and orbit-style maneuvers. For example, it underperforms OpenVLA-UAV on \textit{Surround} (48.30\% vs.\ 100.00\% under fixed instructions) and also shows a gap on \textit{Turn} compared with the strongest VLA baseline. We attribute this limitation to the need for sustained lateral or yaw motion while maintaining visual lock on a central object. The imagination module may gradually drift away from the desired orbital path, causing the generated target to leave the field of view and making downstream motion extraction unstable. 


Overall, these quantitative results demonstrate that predictive world modeling provides a robust and geometry-aware planning mechanism for embodied aerial navigation, particularly in scenarios requiring accurate spatial foresight and stable long-horizon execution.

\begin{figure}[!t]
	\centering
	\includegraphics[width=\linewidth]{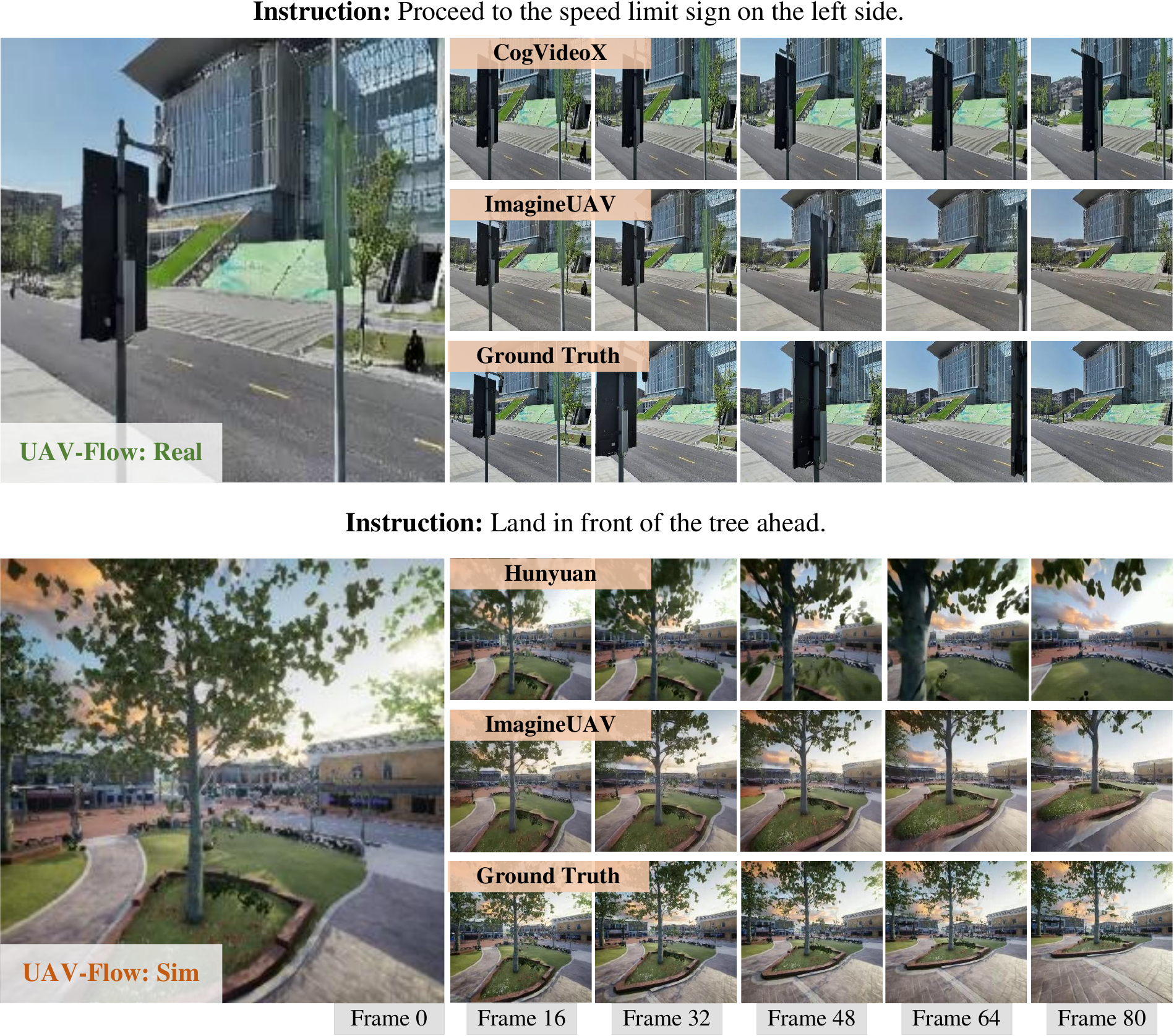}
	\caption{Qualitative comparison of instruction-conditioned visual imagination. Given the same initial observation and language instruction, ImagineUAV is compared with existing video generation backbones on UAV-Flow real and simulated cases, showing that our model generates future egocentric views that better preserve task-directed motion and scene structure.}
	\label{fig:cases}
\end{figure}

\subsection{Qualitative Results}
\subsubsection{Imagination Quality}
We analyze the imagination quality of ImagineUAV through two representative cases in Fig.~\ref{fig:cases}, comparing it with existing video generation backbones under the same initial observation and language instruction. The goal is to evaluate whether the generated future views preserve the task-directed camera motion and scene geometry required for downstream action decoding.

In the real UAV-Flow case, the instruction asks the agent to proceed toward the speed-limit sign from the left side. CogVideoX~\cite{yang2025cogvideox} generates visually plausible frames, but the viewpoint changes only weakly and the sign remains largely static relative to the camera, indicating insufficient motion grounding. In contrast, ImagineUAV produces a clear forward-left viewpoint transition in which the sign moves consistently across the field of view, closely matching the ground-truth rollout and better reflecting the intended navigation behavior.

In the simulated UAV-Flow case, the instruction requires landing in front of the tree. HunyuanVideo-1.5~\cite{wu2025hunyuanvideo} generates future frames with noticeable viewpoint drift and weaker preservation of the target tree geometry. ImagineUAV maintains the tree as the dominant target, preserves the surrounding courtyard structure, and produces a smooth approach sequence that is more consistent with the ground truth. These comparisons show that, compared with generic video generation models, ImagineUAV better converts language intent into temporally coherent egocentric visual futures suitable for geometric action decoding.

\subsubsection{Real-world Experiment}
\begin{figure*}[t]
	\centering
	\includegraphics[width=0.9\linewidth]{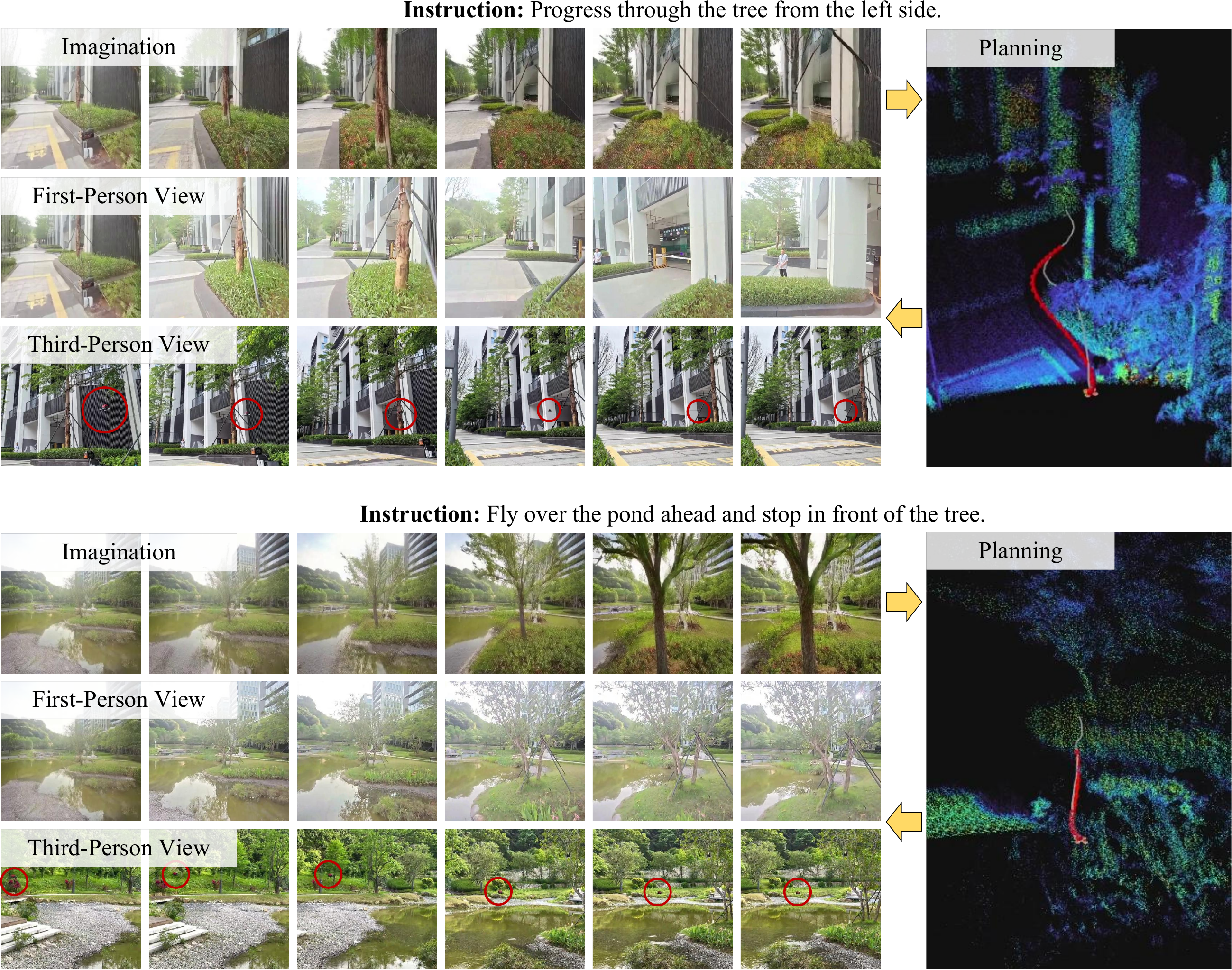}
	\caption{Real-world UAV deployment cases. For each case, the instruction-conditioned world model generates future first-person observations, while the learned action extractor converts the imagined rollout into a planned trajectory executed by the UAV.}
	\label{fig:real_drone}
\end{figure*}

Fig.~\ref{fig:real_drone} presents two real-world deployment cases evaluating the generalization of ImagineUAV in unseen outdoor environments. In the first case, the instruction ``progress through the tree from the left side'' requires obstacle-aware navigation around vegetation. The imagined rollout captures the left-side passing intention, while the action extractor and kinodynamic planner convert it into a smooth and dynamically feasible trajectory for safe execution.

In the second case, the instruction ``fly over the pond ahead and stop in front of the tree'' requires long-range motion and precise stopping behavior. ImagineUAV successfully imagines the viewpoint transition toward the target region, and the kinodynamic planner further refines the extracted trajectory to ensure obstacle clearance and stable flight near the target. These results demonstrate that the proposed imagination-action-planning framework generalizes robustly across different scene layouts, obstacle configurations, and language goals.

\subsection{Ablation Study}

\begin{table}[t]
	\centering
	\caption{Ablation study on uav-flow-sim and real flight.}
	\label{tab:ablation}
	\setlength{\tabcolsep}{2.5pt}
	\small
	\begin{tabular}{lccccc|cc}
		\toprule
		& (a) & (b) & (c) & (d) & (e) & (f) & (g) \\
		\midrule
		\begin{tabular}[c]{@{}l@{}}Imagination\\Module\end{tabular} & $\times$ & $\checkmark$ & $\checkmark$ & $\checkmark$ & $\checkmark$ & $\checkmark$ & $\checkmark$ \\[1pt]
		\addlinespace[2pt]
		\begin{tabular}[c]{@{}l@{}}Action\\Extractor\end{tabular} & $\checkmark$ & $\times$ & $\checkmark$ & $\checkmark$ & $\checkmark$ & $\checkmark$ & $\checkmark$ \\[1pt]
		\addlinespace[2pt]
		\begin{tabular}[c]{@{}l@{}}Kinodynamic\\Planner\end{tabular} & $\checkmark$ & $\checkmark$ & $\times$ & $\checkmark$ & $\checkmark$ & $\checkmark$ & $\times$ \\[1pt]
		\addlinespace[2pt]
		\begin{tabular}[c]{@{}l@{}}10-Step\\Distillation\end{tabular} & $\times$ & $\times$ & $\times$ & $\times$ & $\checkmark$ & $\checkmark$ & $\checkmark$ \\
		\midrule
		Success Rate & 40.2\% & 64.5\% & 68.7\% & 70.9\% & 68.9\% & 13/20 & 9/20 \\
		Inference Time & 14.7s & 11.7s & 14.7s & 14.7s & 6.2s & 6.2s & 6.2s \\
		\bottomrule
	\end{tabular}\\
	{\scriptsize (a)--(e): UAV-Flow-Sim evaluation.\\ 
(f)--(g): real-world flight experiments with 10 navigation tasks and 2 trials per task.}
\end{table}

\begin{table}[t]
	\centering
	\caption{Action Extraction Accuracy.}
	\label{tab:vo_pose}
	\setlength{\tabcolsep}{5pt}
	\small
	\begin{tabular}{lccc}
		\toprule
		& ATE (m) & RPE-t (m) & RPE-r ($^\circ$) \\
		\midrule
		Simulation & 0.098 & 0.142 & 0.715 \\
		Real-world & 0.120 & 0.167 & 0.826 \\
		\bottomrule
	\end{tabular}\\
    {\scriptsize Simulation in Unreal Engine; real-world in a motion capture system.}
\end{table}

\begin{table}[t]
	\centering
	\caption{Computational cost of the World-action pipeline.}
	\label{tab:compute}
	\setlength{\tabcolsep}{4pt}
	\small
	\begin{tabular}{llcc}
		\toprule
		 & Metric & PRO 6000 & AGX Thor \\
		\midrule
		\multirow{3}{*}{\shortstack[l]{Imagination\\Module}} & Inference time & 3.2\,s & 11.7\,s \\
		 & Video duration & 5.4\,s @ 15\,FPS & 5.4\,s @ 15\,FPS \\
		 & VRAM & 18.0\,GB & 18.2\,GB \\
		\midrule
		\multirow{2}{*}{\shortstack[l]{Action\\Extractor}} & Extraction time & 3.0\,s & 9.0\,s \\
		& VRAM & 4.7\,GB & 4.8\,GB \\
		\bottomrule
	\end{tabular}\\
    {\footnotesize The imagination module (10-step distilled) generates 256$\times$256 videos.}
\end{table}

Table~\ref{tab:ablation} reports success rate and inference latency under different module combinations.

\textbf{Effect of the Imagination Module.}
When the imagination module is replaced by the base world model without instruction conditioning and the other modules are kept unchanged, removing instruction-conditioned imagination reduces the success rate from $70.9\%$ to $40.2\%$. This large degradation indicates that the original world model cannot reliably translate language intent into executable 6-DoF egocentric visual rollouts, confirming the importance of motion-aware post-training.

\textbf{Robustness of the Action Extraction.}
Replacing the action extractor with a frame-pair VO baseline decreases the success rate from $70.9\%$ to $64.5\%$, showing that multi-frame motion reasoning is critical for decoding generated videos with temporal artifacts and pixel jitter. Quantitative results in Table~\ref{tab:vo_pose} further demonstrate that the learned action extraction module accurately recovers 6-DoF motion on both simulated and real-world sequences, providing reliable geometric guidance for trajectory generation.

\textbf{Effect of the Kinodynamic Planner.}
Removing the kinodynamic planner and directly executing extracted waypoints reduces the success rate from $70.9\%$ to $68.7\%$ in simulation, but from $13/20$ to $9/20$ in real-world experiments. This indicates that the planner plays a limited role in idealized simulation but is important for bridging the sim-to-real gap by improving robustness against control noise, environmental uncertainty, and dynamically infeasible motions during deployment.

\textbf{Real-world Failure Analysis.}
The full pipeline achieves 13/20 success (Table~\ref{tab:ablation}(f)), with the failures primarily concentrated in yaw-dominant and orbit-style maneuvers—the same difficulty observed in simulation. Specifically, under these instructions the extracted trajectory tends to pass through out-of-view obstacles, and the kinodynamic planner, while ensuring safety, forces the executed motion to deviate from the language instruction. When the kinodynamic planner is disabled (Table~\ref{tab:ablation}(g), 9/20), two primary failure modes emerge: (i) the UAV collides with obstacles mispredicted by the world model, often caused by viewpoint occlusion during sampling or spatiotemporal inconsistency in fine-grained obstacle synthesis; and (ii) the physical response of the UAV cannot track the dynamically unconstrained reference trajectory, leading to instruction deviation or collision.

\textbf{Efficiency of the Step Distillation.}
With all above modules enabled, step-distilled inference reduces the planning time from $14.7$s to $6.2$s, closely approaching the temporal horizon of the generated future observations ($5.4s$). The distilled model maintains a comparable success rate ($68.9\%$ vs. $70.9\%$), demonstrating a favorable trade-off between inference efficiency and navigation performance.

\vspace{-2mm}

\subsection{Discussion}
The results suggest that ImagineUAV's effectiveness mainly comes from generating geometrically meaningful visual foresight rather than photorealistic prediction, providing reliable motion direction and spatial extent cues even with local artifacts. The modular pipeline combining semantic foresight, geometric motion recovery, and kinodynamic planning enables reliable trajectory generation under complex UAV dynamics and obstacle-constrained environments. Computational cost results (Table~\ref{tab:compute}) further suggest the potential for practical onboard deployment, as the current Jetson AGX Thor implementation already approaches the temporal horizon of the generated future observations even without TensorRT acceleration or pipeline parallelism.

\section{Conclusion}
We proposed ImagineUAV, an imagination-driven framework for vision-language navigation for UAVs that bridges high-level language intent and executable flight through predictive visual world modeling. By generating instruction-conditioned future egocentric observations and converting them into executable trajectories via a learning-based action extractor and a kinodynamic planner, ImagineUAV avoids brittle direct action execution and enables stable navigation adhering to the instruction. Extensive simulation and real-world experiments demonstrate state-of-the-art performance on the open-source benchmark and validate real-UAV deployment with step-distilled inference, highlighting the promise of world-model-based predictive planning for embodied aerial navigation. Building on this foundation, our next steps involve migrating the pipeline to onboard hardware to improve real-time performance and incorporating closed-loop feedback for robust long-range navigation.



\bibliographystyle{IEEEtran}
\bibliography{ref}

\vfill
	
\end{document}